\documentclass[10pt,twocolumn,letterpaper]{article}

\usepackage{iccv}
\usepackage{times}
\usepackage{epsfig}
\usepackage{graphicx}
\usepackage{amsmath}
\usepackage{amssymb}
\usepackage{xcolor}
\usepackage{multirow,tabularx}
\usepackage[percent]{overpic}
\usepackage{gensymb}

\usepackage[pagebackref=true,breaklinks=true,letterpaper=true,colorlinks,bookmarks=false]{hyperref}

\renewcommand{\vec}[1]{\boldsymbol{#1}}

\newcommand{\pose}[0]{\vec{\theta}}
\newcommand{\shape}[0]{\vec{\beta}}

\iccvfinalcopy %

\def\iccvPaperID{1288} %

\ificcvfinal\fi
\begin{document}

\title{Tex2Shape: Detailed Full Human Body Geometry From a Single Image}

\author{Thiemo Alldieck\textsuperscript{1,2}
	\and
	Gerard Pons-Moll\textsuperscript{2}
	\and
	Christian Theobalt\textsuperscript{2}
	\and
	Marcus Magnor\textsuperscript{1}
}

\makeatletter
\let\@oldmaketitle\@maketitle%
\renewcommand{\@maketitle}{
    \@oldmaketitle%
    \centering
    \vspace{-4mm}
    \ificcvfinal
    {\small \textsuperscript{1}Computer Graphics Lab, TU Braunschweig, Germany}\\
    {\small	\textsuperscript{2}Max Planck Institute for Informatics, Saarland Informatics Campus, Germany}\\
    {\tt\scriptsize \{alldieck,magnor\}@cg.cs.tu-bs.de \{gpons,theobalt\}@mpi-inf.mpg.de}\\
    \vspace{4mm}
    \fi
    \centering
    \includegraphics[width=0.68\textwidth]{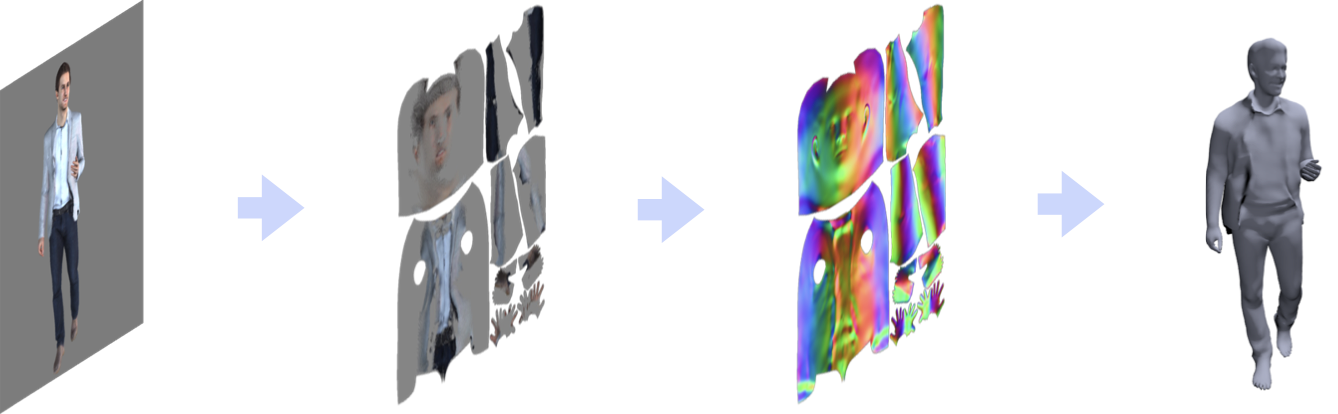}\\
    \refstepcounter{figure}\small Figure~\thefigure: We present an image-to-image translation model for detailed full human body geometry reconstruction from a single image. An input image is transformed into an incomplete texture, then our Tex2Shape network translates the texture into normal and displacement maps. The maps augment a smooth body model with details, hair, and clothing. Result visualized in ground truth pose.
    \label{fig:teaser}
    \vspace{6mm}
}
\makeatother

\maketitle

\thispagestyle{plain}
\pagestyle{plain}

\begin{abstract}
\vspace{-2mm}
   We present a simple yet effective method to infer detailed full human body shape from only a single photograph.
   Our model can infer full-body shape including face, hair, and clothing including wrinkles at interactive frame-rates.
   Results feature details even on parts that are occluded in the input image.
   Our main idea is to turn shape regression into an aligned image-to-image translation problem.
   The input to our method is a partial texture map of the visible region obtained from off-the-shelf methods.
   From a partial texture, we estimate detailed normal and vector displacement maps, which can be applied to a low-resolution smooth body model to add detail and clothing. 
   Despite being trained purely with synthetic data, our model generalizes well to real-world photographs.
   Numerous results demonstrate the versatility and robustness of our method. 
\end{abstract}

\vspace{-3mm}
\section{Introduction}
\label{sec:introduction}

In this paper, we address the problem of automatic \mbox{\emph{detailed}} full-body human shape reconstruction from a single image.
Human shape reconstruction has many applications in virtual and augmented reality, scene analysis, and virtual try-on.
For most applications, acquisition should be quick and easy, and visual fidelity is important.
Reconstructed geometry is most useful if it shows hair, face, and clothing folds and wrinkles at sufficient detail -- what we refer to as detailed shape.
Detail adds realism, allows people to feel identified with their self-avatar and their interlocutors, and often carries crucial information.

While a large number of papers focus on recovering pose, and rough body shape from a single image~\cite{omran2018neural,kanazawa2018endtoend,pavlakos2018humanshape,bogo2016smplify}, much fewer papers focus on recovering detailed shapes. 
Some recent methods recover pose and non-rigid deformation from monocular video~\cite{MonoPerfCap_SIGGRAPH2018}, even in real-time~\cite{Habermann:2019:LiveCap}. However, they require a pre-captured static template of each subject.   
Other recent works~\cite{alldieck2018video,alldieck2019learning} recover static body shape, and clothing as displacements on top of the SMPL body model~\cite{smpl2015loper} (model-based), or use a voxel representation~\cite{varol17_surreal,natsume2018siclope}. 
Voxel-based methods~\cite{varol17_surreal,natsume2018siclope} often produce errors at the limbs of the body and require fitting a model post-hoc~\cite{varol17_surreal}.
Model-based methods are more robust, but results tend to lack fine detail. We hypothesize there are three reasons for this. 
Firstly, they rely mostly on silhouettes for either fitting~\cite{alldieck2018video}, or CNN-based regression plus fitting~\cite{alldieck2019learning}, ignoring the rich illumination and shading information contained in RGB values. 
Secondly, the regression from image pixels directly to 
3D mesh displacements 
is hard because inputs and outputs are \emph{not aligned}.
Furthermore, prediction of high-resolution meshes requires mesh-based neural networks, which are very promising but are harder to train than standard 2D CNNs. 
Finally, they rely on 3D pose estimation, which is hard to obtain accurately. 

Based on these observations, our idea is to turn the shape regression into an \emph{aligned} image-to-image translation problem (see Fig.~\ref{fig:teaser}).
To that end, we map input and output pairs to the pose-independent UV-mapping of the SMPL model.
The UV-mapping unfolds the body surface onto a 2D image such that every pixel corresponds to a 3D point on the body surface.
Similar to \cite{neverova2018dense}, we map the visible image pixels to the UV space using DensePose~\cite{alp2018densepose} obtaining a partial texture map image, which we use as input.
Instead of regressing details directly on the mesh, we propose to regress shape as UV-space displacement and normal maps.
Every pixel stores a normal and a displacement vector from a smooth shape (in the space of SMPL) to the detailed shape.
We call our model to Tex2Shape.

We train Tex2Shape with a dataset of $2043$ 3D scans of people in varying clothing, poses, and shapes. 
To map all scan shapes to the UV-space, we non-rigidly register SMPL to each scan, optimizing for model shape parameters and free-form displacements, and store the latter in a displacement map.
Registration is also useful for augmentation; using SMPL, we render multiple images of varying pose and camera view.
We further augment the renderings with realistic illumination, which is a strong cue in this problem.
Assuming a Lambertian reflectance model, we know that color forms from the dot product of light direction and the surface normal times albedo.
Shape-from-shading~\cite{zhang1999shape} allows to invert the process and estimate the surface from shading, which was used before to refine geometry of stereo-based~\cite{wu2013set} or multi-view-based human performance capture results~\cite{Wu:2012,LWSLVDT13}.
After synthesizing image pairs, we train a Pix2Pix network~\cite{isola2017pix2pix} to map from partial texture maps to complete normal and displacement maps and a second small network for estimating SMPL body shape parameters.

Several experiments demonstrate that our proposed data pre-processing undoubtedly pays-off. Trained only from synthetic images, our model can robustly produce, in one shot, \emph{full 3D shapes} of people with varied clothing, shape, and hair. In contrast to models that produce normals or shading only for the visible image part, Tex2Shape hallucinates the shape also for the \emph{occluded} part -- effectively performing translation and completion together.
In summary, our contributions are: 
\vspace{-.5mm}
\begin{itemize}
    \item We turn a hard full-body shape reconstruction problem into an easier 3D pose-independent image-to-image translation one. To the best of our knowledge, this is the first method to infer detailed body shape as image-to-image translation. \vspace{-1.8mm}
    \item From a single image, our model 
    can regress full 3D clothing, hair and facial details in $50$~milliseconds. \vspace{-1.8mm}
    \item Experiments demonstrate that, while very simple, Tex2Shape is very effective and is capable of regressing full 3D clothing, hair and facial details in a static reference pose in one shot.  \vspace{-1.8mm}
    \item Tex2Shape is available for research purposes~\cite{code}.
\end{itemize}

\section{Related Work}
\label{sec:related}
Human shape reconstruction is a wide field of research, often jointly approached with pose reconstruction.
In the following, we review methods for human pose and shape reconstruction from monocular image and video.
Full body methods are often inspired by methods for face geometry estimation.
Hence, we include face reconstruction in our review.
When it comes to detailed reconstruction, clothing plays an important role.
Therefore, we conclude with a brief overview of garment reconstruction and modeling.

\vspace{-4mm}
\paragraph{Pose and shape reconstruction.}
Methods for monocular pose and shape reconstruction often utilize parametric body models to limit the search space~\cite{anguelov2005scape,hasler2009statistical,smpl2015loper,pons2015dyna,joo2018total}, or use a pre-scanned static template to capture pose and non-rigid surface deformation~\cite{MonoPerfCap_SIGGRAPH2018,Habermann:2019:LiveCap}.
To recover pose and shape, the 3D body model is fitted to 2D poses.
In early works 2D poses have been entirely or partially manually clicked~\cite{guan2009estimating,zhou2010parametric,jain2010moviereshape,rogge2014garment},
later the process was automated~\cite{bogo2016smplify,Lassner} with 2D landmark detections from deep neural networks~\cite{pishchulin16cvpr,insafutdinov2016deepercut,cao2017realtime}.
In recent work, the SMPL~\cite{smpl2015loper} model has been integrated into network architectures~\cite{kanazawa2018endtoend,pavlakos2018humanshape,omran2018neural,tung2017self}. This further automates and robustifies the process.
All these works focus mostly on robust pose detection. Shape estimation is often limited to surface correlations with bone lengths. Most importantly, the shape is limited to the model space.
In contrast, we focus only on shape and estimate geometry details beyond the model space.

Clothing and hair can be obtained by optimization-based methods \cite{alldieck2018video,alldieck2018detailed}.
From a video of a subject turning around in A-pose, silhouettes are fused in canonical pose.
In the same setting, the authors in \cite{alldieck2019learning} present a hybrid learning and optimization-based method, that makes the process completely automatic, fast, and dependent only on a handful of images.
However, all these methods can only process A-poses and depend on robust pose detection. The method in~\cite{weng2018photo} loosens this restriction and creates humanoid shapes from a single image via 2D warping of SMPL parameters, but only partially handles self-occlusion. Another recent line of research estimates pose and shape in form of a voxel representation \cite{varol2018bodynet,jackson20183d,natsume2018siclope}, which allows for more complex clothing but limits the level of detail. In \cite{Zheng2019DeepHuman} the authors alleviate this limitation by augmenting the visible parts with a predicted normal map. In contrast, we present 3D pose-independent shape estimation in a reference pose with high-resolution details also on non-visible parts.

Several previous methods exploited shading cues in high-frequency texture to estimate high-frequency detail. For instance, they estimated lighting and reflectance to compute shape-from-shading-refined geometry of a human template from stereo~\cite{wu2013set} or multi-view imagery~\cite{Wu:2012,LWSLVDT13}.

\vspace{-4mm}
\paragraph{Face reconstruction.}
Several recent monocular face reconstruction and performance capture methods use shading-based refinement for geometry improvement, e.g., in analysis-by-synthesis fitting~\cite{sela2017unrestricted} or refinement, or in a trained neural network~\cite{sengupta2018sfsnet,huynh2018mesoscopic}. 
Also related to our approach are recent works integrating a differentiable face renderer in a neural network to estimate instance correctives of geometry and albedo relative to a base model~\cite{tewari2018self}, or learn an identity geometry and albedo basis from scratch from video~\cite{FML2019}. 

\vspace{-4mm}
\paragraph{Garment reconstruction and modeling.}
Body shape under clothing has been estimated without~\cite{zahng2017shapeundercloth} and jointly with a separate clothing layer~\cite{ponsmoll2017clothcap} from 3D scans and from RGB-D~\cite{SimulCap19}. \cite{yang2018analyzing} introduces a technique, which allows complex clothing to be modeled as offsets from the naked body.
The work in~\cite{wang2018learning} describes a model that encodes shape, garment sketch, and garment model, in a single shared latent code, which enables interactive garment design. 
High frequency wrinkles are predicted as a function of pose either in UV space using a CNN~\cite{lahner2018deepwrinkles,jin2018pixel} or directly in 3D using a data-driven optimization method~\cite{popa2009wrinkling}. All these methods~\cite{lahner2018deepwrinkles,yang2018analyzing,jin2018pixel} target realistic \emph{animation} of clothing and can only predict garments in isolation~\cite{lahner2018deepwrinkles,jin2018pixel}.
Learning based normals and depth recovery~\cite{bednarik2018learning} or meshes~\cite{danvevrek2017deepgarment} has been demonstrated but again only for single garments. 
In contrast, our approach is the first to reconstruct the detailed shape of a \emph{full-body} from a \emph{single image} by learning an image-to-image mapping.

\section{Method}
\label{sec:method}
\begin{figure*}
    \centering
    \includegraphics[width=\textwidth]{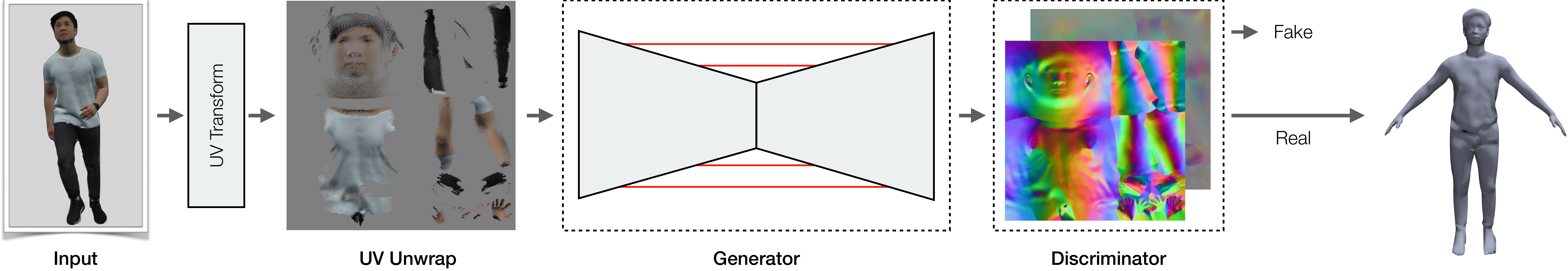}
    \caption{Overview of the key component of our method: A single photograph of a subject is transformed into a partial UV texture map. This map is then processed with a U-Net with skip connections that preserve high-frequent details. A PatchGAN discriminator enforces realism. The generated normals and displacements can be applied to the SMPL model using standard rendering pipelines.}
    \label{fig:overview}
    \vspace{-2.5mm}
\end{figure*}

The goal of this work is to create an animatable 3D model of a subject from a single photograph.
The model should reflect the subject's body shape and contain details such as hair and clothing with garment wrinkles.
Details should be present also on body parts that have not been visible in the input image, e.g.\ on the back of the person.
In contrast to previous work \cite{natsume2018siclope,weng2018photo,alldieck2019learning} we aim for fully automatic reconstruction which does not require accurate 3D pose. 
To this end, we train a Pix2Pix-style~\cite{isola2017pix2pix} convolutional neural network to infer normals and vector displacement (\emph{UV shape-images}) on top of the SMPL body model~\cite{smpl2015loper}. To align the input image with the output UV-shape images, we extract a partial UV texture map of the visible area using off-the-shelf methods~\cite{alp2018densepose,kanazawa2018endtoend}.
An overview is given in Fig.~\ref{fig:overview}.
A second small CNN infers SMPL shape parameters from the image (see Sec.~\ref{sec:architecture}).
In Sec.~\ref{sec:smpl} we describe the parametric body model used in this work, and in Sec.~\ref{sec:uv} we explain our parameterization of appearance, normals, and displacements.

\subsection{Parametric body model}
\label{sec:smpl}
SMPL is a parameterized body model learned from scans of subjects in minimal clothing.
It is defined as a function of pose $\pose$ and shape $\shape$ returning a mesh of $N = 6890$ vertices and $F = 13776$ faces.
Shape $\shape$ corresponds to the first $10$ principal components of the training data subjects.
Since scale is an inherent ambiguity in monocular images, we made $\shape$ independent of body height in this work.
Our method estimates $\shape$ with a standardized height and is independent of pose $\pose$.
Details that go beyond the SMPL shape space are added via UV displacement and normal maps (UV shape-images), as described in Sec.~\ref{sec:uv}.
During the dataset generation (see Sec.~\ref{sec:dataset}), we use SMPL to synthesize images of humans posing in front of the camera.

\subsection{UV parameterization}
\label{sec:uv}
The SMPL model describes body shapes with a mesh containing $6890$ vertices.
Unfortunately, this resolution is not high enough to explain fine details, such as garment wrinkles.
Another problem is that meshes do not live on a regular 2D grid like images, and consequently require taylored solutions~\cite{bronstein2017geometric} that are not yet as effective as standard CNNs on the image domain. 
To leverage the power of standard CNNs, we propose to use a well-established parameterization of mesh surfaces: UV mapping~\cite{blinn1976texture}.
A UV map unwraps the surface onto an image, allowing to represent functions defined on the surface as images. Hereby, $U$ and $V$ denote the 2 axes of the image.
The mapping is defined once per mesh topology and assigns every pixel in the map to a point on the surface via barycentric interpolation of neighboring vertices.
By using a UV map, a mesh can be augmented with geometric details of a resolution proportional to the UV map resolution.

We augment SMPL using two UV maps, namely normal map and vector displacement map.
A normal map contains new surface normals, that can add or enhance visual details through shading.
A vector displacement map contains 3D vectors that displace the underlying surface. Displacements and normals are defined on the canonical T-pose of SMPL.
The input to our neural network is a partial texture map of the visible pixels on the input photograph (see Sec.~\ref{sec:create_partial_tex}).

\section{Dataset Generation}
\label{sec:dataset}
To learn our model we synthesize a varied dataset from real 3D scans of people. Specifically, we synthesize images of humans in various poses under realistic illumination paired with normal maps, displacement maps, and SMPL shape parameters $\shape$. The large majority of scans (1826) was kindly provided from Twindom (https://web.twindom.com/). We additionally purchased 163 scans from renderpeople.com and 54 from axyz-design.com.
These scans do not share the same mesh layout, and therefore we can not directly compute coherent normal and displacement maps. To this end, we non-rigidly register the SMPL model against each of the scans.
This ensures that all vertices share the same contextual information across the dataset.
Furthermore, we can change the pose of the scans using SMPL.
Unfortunately, non-rigid registration of clothed people is a very challenging problem itself (see Sec.~\ref{sec:registration}), and often results in unnatural shapes.
Hence, we manually selected 2043 high quality registrations. Unfortunately, our current dataset is slightly biased towards men because registration currently fails more often for women, due to long hair, skirts and dresses. 
Of the 2043 scans, we reserve 20 scans for validation and 55 scans for testing.

In the following, we explain our non-rigid registration procedure in more detail and describe the synthetization of the paired dataset for training of the models.

\subsection{Scan registration}
\label{sec:registration}
As discussed in Sec.~\ref{sec:smpl}, $N=6890$ vertices are not enough to explain fine details.
To this end, we sub-divide each face in SMPL into four, resulting in a new mesh consisting of $N=27554$ vertices and $F=55104$ faces.
This high-resolution mesh can better explain fine geometric details in the scans.
While joint optimization is generally desirable, registration is much more robust when done in stages: we first compute 3D pose, then body shape and finally non-rigid details. We start the registration by reconstructing the pose of the scan subject.
Therefore, we find 3D landmarks by rendering the scan from multiple cameras and minimizing the 2D re-projection error to 2D joint OpenPose detections~\cite{cao2017realtime}.
Then we optimize the SMPL pose parameters $\pose$ to explain the estimated 3D joint locations.
Next, we optimize for shape parameters $\shape$ to minimize scan to SMPL surface distance.
Here, we make sure SMPL vertices stay inside the scan by paying a higher cost for vertices outside the scan since SMPL can only reliable explain the naked body shape.
Finally, we recover fine-grained details by optimizing the location of SMPL vertices.
The resulting registrations explain high-frequency details of the scans with the subdivided SMPL mesh layout and can be re-posed.

\subsection{Spherical harmonic lighting}
For a paired dataset, we first need to synthesize images of humans.
For realistic illumination, we use spherical harmonic lighting.
Spherical harmonics (SH) are orthogonal basis functions defined over the surface of the sphere.
For rendering SH are used to describe the directions from where light is shining into the scene~\cite{ramamoorthi2001efficient}.
We follow the standard procedure and describe the illumination with the first 9 SH components per color.
To produce a large variety of realistic illumination conditions, we convert images of the \emph{Laval Indoor HDR dataset}~\cite{gardner-sigasia-17} into diffuse SH coefficients, similar to~\cite{kanamori2018relighting}.
For further augmentation, we rotate the coefficients randomly around the Y-axis.

\subsection{UV map synthetization}
To complete our dataset, we calculate UV maps that explain details of the 3D registrations.
In UV mapping every face of the mesh has a 2D counterpart in the UV image.
Hence, UV mapping is essentially defined through a 2D mesh.
Given a 3D mesh and a set of per-vertex information, a UV map can be synthesized through standard rendering.
Information between vertices is filled through barycentric interpolation.
This means, given the high-resolution registrations, we can simply render detailed UV displacement and normal maps.
The displacement maps encode the free-form offsets, that are not part of SMPL.
The normal maps contain surface normals in canonical T-pose.
These maps are used to augment the standard-resolution naked SMPL, which eliminates the need for higher mesh-resolution or per-vertex offsets. We use the standard-resolution SMPL augmented with the UV maps in all our experiments.

\section{Model and Training}
\label{sec:model}
In the following, we explain the used network architectures, losses, and training schemes in more detail.
Further, we explain how a partial texture can be obtained from DensePose~\cite{alp2018densepose} results.

\subsection{Network architectures}
\label{sec:architecture}

Our method consists of two CNNs -- one for normal and displacement maps and one for SMPL shape parameters $\shape$.
The main component of our method is the Tex2Shape-network as depicted in Fig.~\ref{fig:overview}.
The network is a conditional Generative Adversarial Network (Pix2Pix) \cite{isola2017pix2pix} consisting of a U-Net generator and a PatchGAN discriminator.
The U-Net features each seven convolution-ReLU-batchnorm down- and up-sampling layers with skip connections.
The discriminator consists of four of such down-sampling layers.
We condition on $512\times512$ partial textures, based on two observations: First, when mapping pixels from an HD $1024\times1024$ image to UV, the resolution is high enough to contain most pixels from the foreground, and not too high to prevent large unoccupied regions. 
Second, using the mesh resolution of the training set, larger UV maps would only contain more interpolated data. 
See supplemental material for an ablation experiment using smaller UV maps.

The $\shape$-network takes $1024\times1024$ DensePose detections as input.
These are then again down-sampled with seven convolution-ReLU-batchnorm layers and finally mapped to $10$ $\shape$-parameters by a fully-connected layer.

\subsection{Losses and training scheme}
\label{sec:losses}
The goal of our method is to create results with high \emph{perceived quality}.
We believe structure is more important than accuracy and therefore experiment with the following loss:
The structural similarity index (SSIM) was introduced to predict the perceived quality of images.
The multi-scale SSIM (MS-SSIM) \cite{wang2003multiscale} evaluates the image on different image scales.
We maximize the structural similarity of ground truth and predicted normal and displacement maps by minimizing the dissimilarity (MS-DSSIM): $(1-\text{MS-SSIM})/2$. We further train with the well-established L1-loss and the GAN-loss coming from the discriminator. Finally, the $\shape$-network is trained with an L2 parameter loss.
We train both CNNs with the Adam optimizer~\cite{kingma2014adam} and decay the learning-rate once the losses plateau.

\subsection{Input partial texture map}
\label{sec:create_partial_tex}
\begin{figure}
    \centering
    \includegraphics[width=0.82\columnwidth]{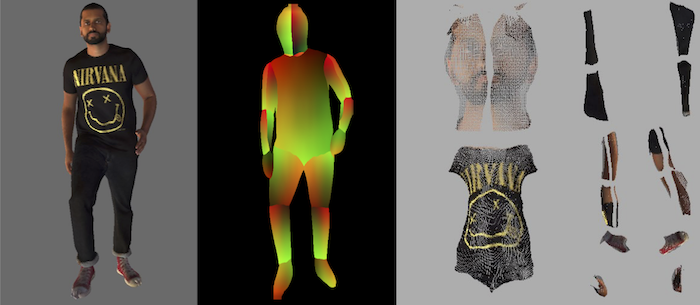}\\
    \caption{To create the input to our method, we first process the input image (left) with DensePose. The DensePose result (middle) contains UV coordinates, that can be used to map the input image into a partial texture (right).}
    \label{fig:uv_mapping_explained}
    \vspace{-4mm}
\end{figure}

The partial texture forming the input to our method is created by transforming pixels from the input image to UV space based on DensePose detections, see Fig.~\ref{fig:uv_mapping_explained}.
DensePose predicts UV coordinates of $24$ body parts of the SMPL body model (Fig.~\ref{fig:uv_mapping_explained} middle).
For easier mapping, we pre-compute a look-up table to convert from $24$ DensePose UV maps to the single joint SMPL UV parameterization.
Each pixel in the DensePose detection now maps to a coordinate in the SMPL UV map.
Using this mapping, we compute a partial texture from the input image (Fig.~\ref{fig:uv_mapping_explained} right).

\section{Experiments}
\label{sec:experiments}

\begin{figure*}
    \centering
    \includegraphics[width=\textwidth]{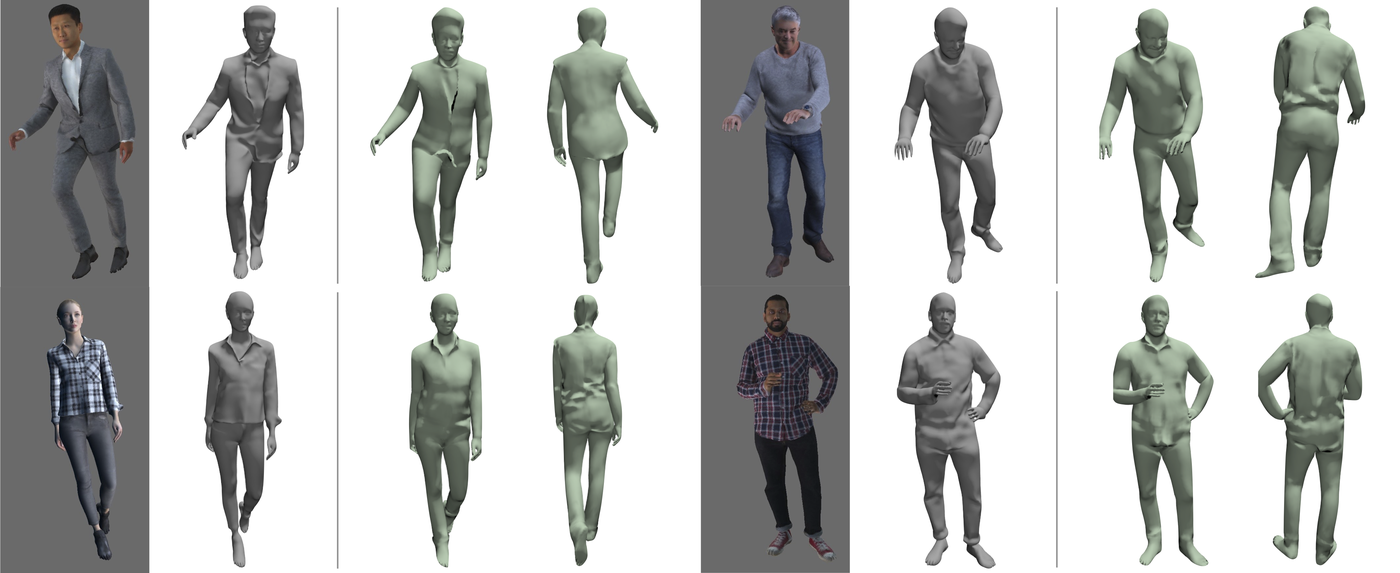}
    \includegraphics[width=\textwidth]{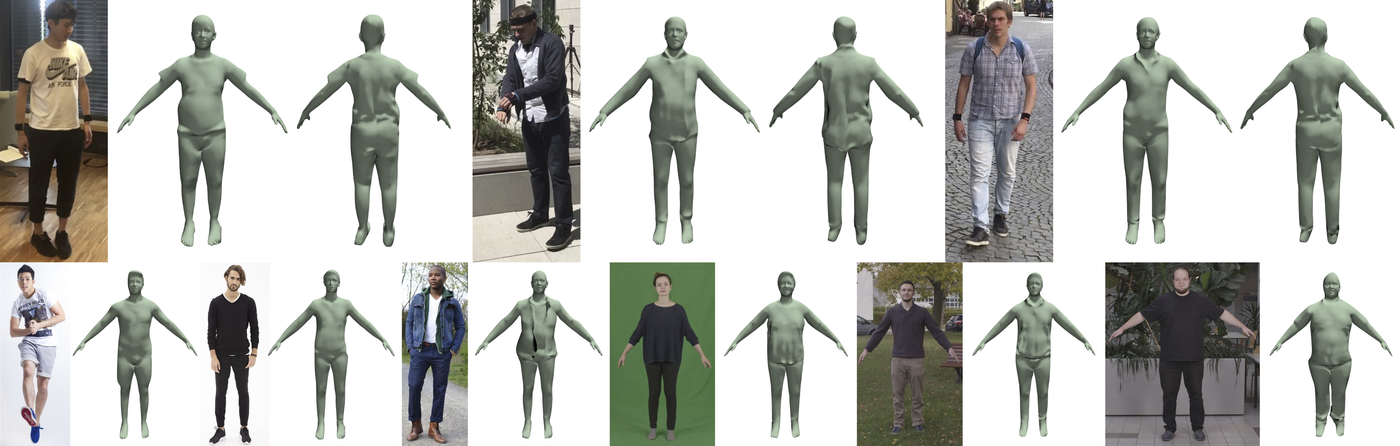}
    \caption{Our 3D reconstruction results (green) on four different datasets. We compare to ground truth (grey) on our synthetic dataset (rows 1 and 2). Qualitative results on 3DPW (3rd row), DeepFashion (4th row left) and PeopleSnapshot (4th row right) demonstrate, that our model generalizes well to real-world footage. Details on the back of the models are hallucinated by our model.}
    \label{fig:qualitative_results}
    \vspace{3mm}
    \centering
    \includegraphics[width=\textwidth]{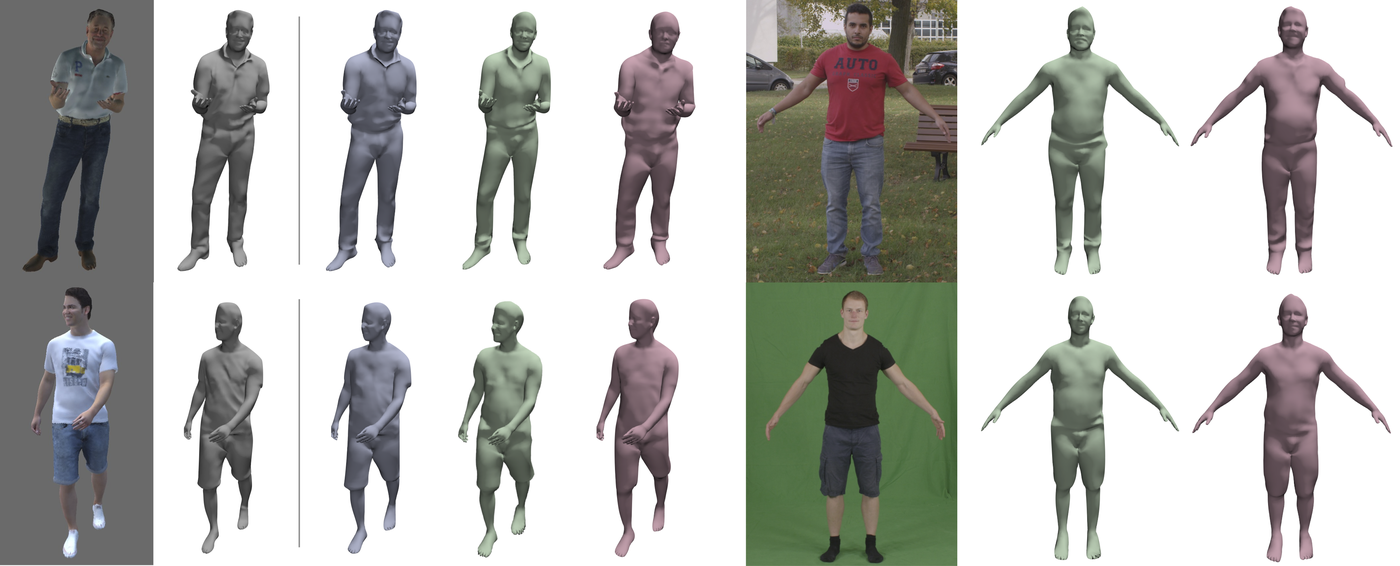}\\
    \caption{Results using three different methods for partial texture creation compared to input image and ground truth mesh (grey): ground truth UV mapping (blue), DensePose UV mapping (green), HMR-based texture reprojection (red), cf.~Fig.~\ref{fig:partial_textures}.}
    \label{fig:uv_mapping_impact}
\end{figure*}

In the following, we qualitatively and quantitatively evaluate our proposed method.
Results on four different datasets and comparisons to state-of-the-art demonstrate the versatility and robustness of our method as well as the quality of results (Sec~\ref{sec:quali_results}).
Further, we study the effect of different supervision losses (Sec.~\ref{sec:type_of_supervision}), evaluate different methods for UV mapping (Sec.~\ref{sec:impact_of_uv_mapping}), and measure the robustness for different visibility levels (Sec.~\ref{sec:impact_of_visibility}).
Finally, in Sec.~\ref{sec:garment_transfer} we demonstrate a potential application of our proposed method, namely garment transfer between subjects.
More experiments and ablation studies can be found in the supplemental material.
Due to scale ambiguity in monocular images, all results are up to scale.
Also, our method does not compute pose.
For better inspection, we depict results in ground truth or A-pose.
Further, we color-code the results by the used method for UV-mapping (see Sec.~\ref{sec:impact_of_uv_mapping}).
Results using DensePose mapping are \emph{green}, \emph{blue} marks ground truth mapping, \emph{red} indicates HMR-based~\cite{kanazawa2018endtoend} texture reprojection, and ground truth shapes are \emph{grey}.
All results have been calculated at interactive frame-rates. Precisely, our method takes on average $50$~ms for displacement map, normal map, and $\shape$-estimation on an NVIDIA Tesla V100. UV mapping using DensePose can be performed in real-time. 

\subsection{Qualitative results and comparisons}
\label{sec:quali_results}

\begin{figure}
    \centering
    \includegraphics[width=0.95\columnwidth]{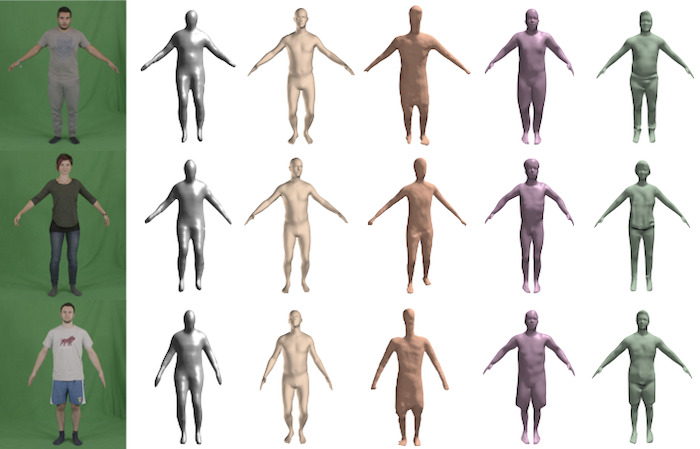}\\
    \caption{Our method compared to other methods for human shape reconstruction. From left to right: Input image, BodyNet~\cite{varol2018bodynet}, HMR~\cite{kanazawa2018endtoend}, SiCloPe~\cite{natsume2018siclope}, Video Shapes~\cite{alldieck2018video}, and ours. Our method preserves the highest level of detail.} 
    \label{fig:qualitative_comparisons}
    \vspace{-3mm}
\end{figure}

We qualitatively compare our work against four relevant methods for monocular human shape reconstruction on the \emph{PeopleSnapshot} dataset \cite{alldieck2018video}.
BodyNet~\cite{varol2018bodynet} is a voxel-based method to estimate human pose and shape from only one image.
SiCloPe~\cite{natsume2018siclope} is voxel-based, too, but recovers certain details by relying on synthesized silhouettes of the subject.
HMR~\cite{kanazawa2018endtoend} is a method to estimate pose and shape from single image using the SMPL body model.
In \cite{alldieck2018video} the authors present the first video-based monocular shape reconstruction method, that goes beyond the parameters of SMPL.
They use 120 images of the same subject roughly posed in A-poses and fuse the silhouettes into a canonical representation.
However, the method is optimization-based and requires to fit the pose in each frame first, which makes the process very slow.
In Fig.~\ref{fig:qualitative_comparisons}, we show a side-by-side comparison with our results.
Our method clearly features the highest level of detail, even compared to \cite{alldieck2018video} using 120 frames, while our method only takes a single image as input and runs at interactive frame-rates.

In Fig.~\ref{fig:qualitative_results} we show more results of our method.
We compare against ground truth on our own dataset and show qualitative results on \emph{3DPW}~\cite{vonMarcard2018}, \emph{DeepFashion}~ \cite{liuLQWTcvpr16DeepFashion,liuYLWTeccv16FashionLandmark}, and \emph{PeopleSnapshot}~\cite{alldieck2018video} datasets.
Our method successfully generalizes to various real-world conditions.
Please note how realistic garment wrinkles are hallucinated on the unseen back of the models.
In general, we can see our method is able to infer realistic 3D models featuring hair, facial details, and various clothing including garment wrinkles from single image inputs.

\subsection{Type of supervision}
\label{sec:type_of_supervision}
\begin{figure}
    \centering
    \includegraphics[width=0.95\columnwidth]{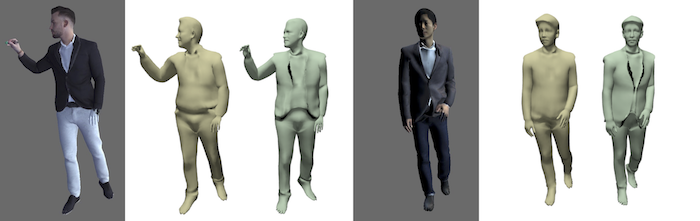}\\
    \caption{After training with MS-DSSIM loss enabled (green) complex clothing is reconstructed more reliably, than after training with L1 loss only (yellow).}
    \label{fig:type_of_supervision}
    \vspace{-3mm}
\end{figure}

In Sec.~\ref{sec:losses}, we have introduced the MS-DSSIM loss.
The intuition behind using this loss is that for visual fidelity structure is more important than accuracy.
To evaluate this design decision, we train a variant of our Tex2Shape network with L1 and GAN losses only.
Since it is not straight forward to quantify better structure, we closely inspect our results on a visual basis.
We find, that the variant trained with MS-DSSIM loss is able to reconstruct complex clothing more reliably. Examples are shown in Fig.~\ref{fig:type_of_supervision}.
Note that the results computed with MS-DSSIM loss successfully reconstruct the jackets.

\subsection{Impact of UV mapping}
\label{sec:impact_of_uv_mapping}
\begin{figure}
    \centering
    \includegraphics[width=0.95\columnwidth]{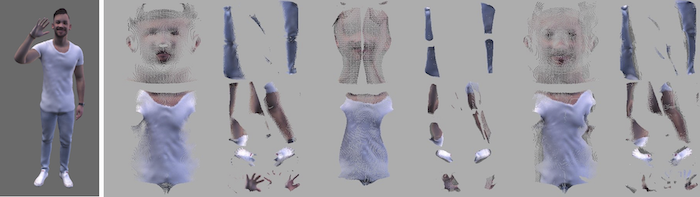}\\
    \caption{Partial textures computed with different methods. From left to right: Input, ground truth UV mapping, DensePose, HMR.}
    \label{fig:partial_textures}
    \vspace{-1mm}
\end{figure}

Our method requires to first map an input image to a partial UV texture.
We propose to use DensePose~\cite{alp2018densepose}, which makes our method independent of the 3D pose of the subject.
In the following, we evaluate the impact of the choice of UV mapping on our method.
To this end, we train three variants of our network.
Firstly, we train with ground truth UV mappings calculated from the scans. We render the scan's UV coordinates in image space, that are then used for UV mapping, similar to the mapping using DensePose (see Sec.~\ref{sec:create_partial_tex}). We refer to this variant as \emph{GT-UV}.
Secondly, we train a variant that can be used with off-the-shelf 3D pose estimators.
To this end, we render UV coordinates of the naked SMPL model without free-form offsets. This way only pixels that are covered by the naked SMPL shape are mapped, what simulates texture reprojection from results of 3D pose detectors (\emph{3D pose variant}).
Finally, we compare with our standard training procedure using DensePose.
A comparison of partial textures created with the three variants is given in Fig.~\ref{fig:partial_textures}.
Note how we lose large parts of the texture by using DensePose mapping.

To evaluate the 3D pose variant, we choose HMR~\cite{kanazawa2018endtoend} as 3D pose detector.
Unfortunately, the results of HMR do not always align with the input image what produces large errors in the UV space.
To this end, we refine the results by minimizing the 2D reprojection error of SMPL joints to OpenPose~\cite{cao2017realtime} detections. We choose dogleg optimization and optimize for $20$ steps.

In Fig.~\ref{fig:uv_mapping_impact} we show a side-by-side comparison of the three variants.
While GT-UV and DensePose variants are almost identical, the 3D pose variant lacks some detail and introduces noise in the facial region.
This is caused by the fact, that perfect alignment is still not achieved even after pose-refinement.
The GT-UV and DensePose variants differ the most in hairstyle and at the boundary of the shorts, what is not surprising since hair and clothing are only partially mapped by DensePose.
However, both variants closely resemble ground truth results.
The DensePose and 3D pose mapping variants can directly be used on real-world footage, while only being trained with synthetic data.

\subsection{Impact of visibility}
\label{sec:impact_of_visibility}


In the following, we numerically evaluate the robustness of our method to different visibility settings caused by different poses and distances to the camera.
The following results have been computed using GT UV mapping to factor out noise introduced by DensePose.
Which pixels can be mapped to the UV partial texture is determined by the subject's pose and distance to the camera.
Parts of the body might be not visible (e.g.\ the subject's back) or occluded by other body parts.
If the subject is far away from the camera, it only covers only a small area of the image and thus only a small number of pixels can be mapped.

\begin{figure}
    \centering
    \includegraphics[width=0.95\columnwidth]{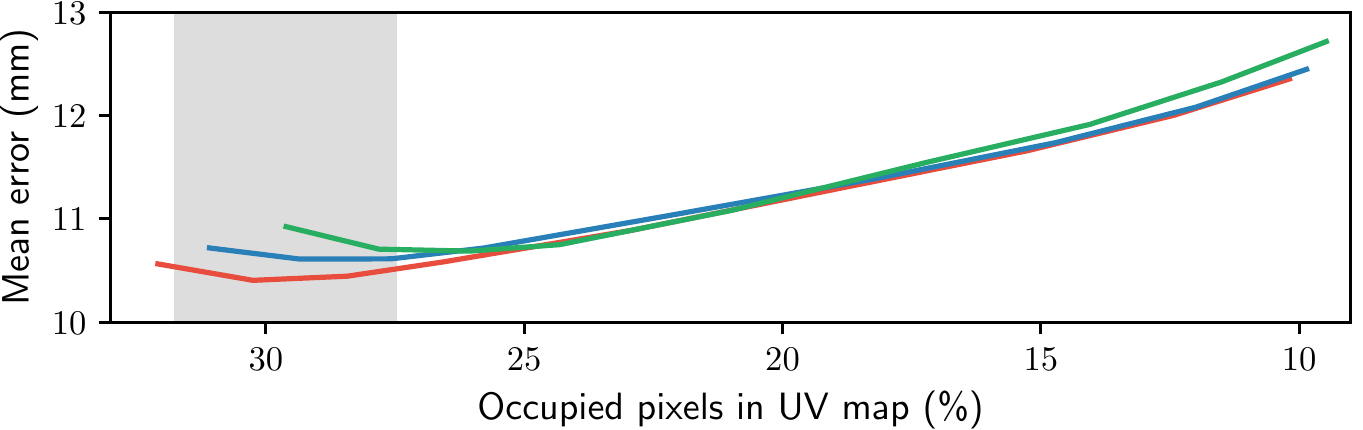}\\
    \caption{Average displacement error for three different poses (red: A-pose, blue: walking, green: posing sideways with hands touching) and different distances to the camera. The shaded region marks the margin of trained UV map occupancy.}
    \label{fig:occlusion_vs_accuracy}
    \vspace{2mm}
    \centering
    \includegraphics[width=0.95\columnwidth]{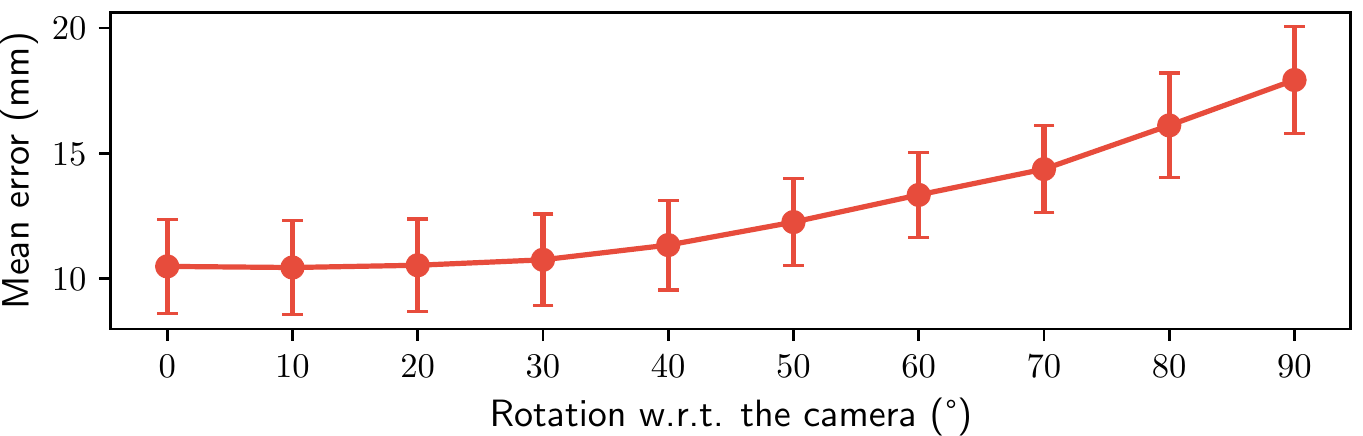}\\
    \caption{Average displacement error for A-posed subjects and different rotations around Y-axis with respect to the camera. Our model has been trained on rotations $\pm20\degree$.}
    \label{fig:rotation_vs_accuracy}
    \vspace{-2mm}
\end{figure}

In Fig.~\ref{fig:occlusion_vs_accuracy} we measure how this influences the accuracy of our results.
Over a test-set with $55$ subjects, we synthesize images of three different poses with various distances to the camera.
The three poses are A-pose, walking towards the camera, and posing sideways with hands touching.
We report the mean per-pixel error of 3D displacements maps (including unseen areas) against the percentage of occupied pixels in the partial texture.
For all three poses, the error increases linearly, even for untrained texture occupations.
Not surprisingly, the minimum of all three poses lies in the margin of trained occupations.
Admittedly, for higher occupations, the error slightly goes up what is caused by the fact, that the network was not trained for scenarios where the subject fully covers the input image.

In Fig.~\ref{fig:rotation_vs_accuracy}, we study the robustness of our method against unseen poses.
We trained the network with images of humans roughly facing the camera.
Therefore, we randomly sampled poses in our dataset and Y-axis rotations between $\pm20\degree$.
In this experiment, we rotate an A-pose around the Y-axis and report the mean per-pixel 3D displacement error.
From $0\degree$ to $30\degree$, the error stays almost identical, after $30\degree$ it increases linearly.
Again this behavior can be explained by the network not being trained for such angles.

Both experiments demonstrate the robustness of our method against scenarios not covered by our training set.

\subsection{Garment transfer}
\label{sec:garment_transfer}

\begin{figure}
    \centering
    \includegraphics[width=0.55\columnwidth]{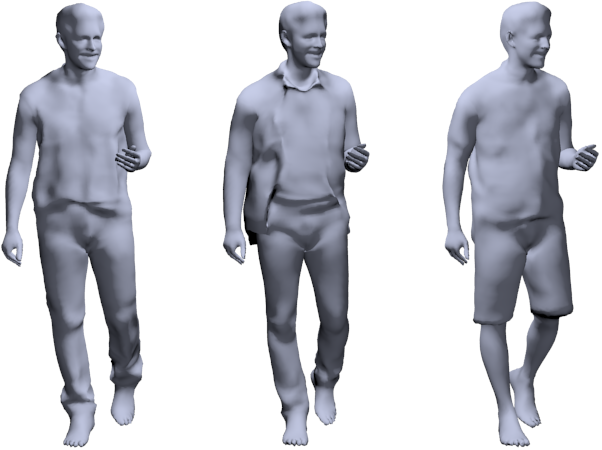}\\
    \caption{Since all reconstructions share the same mesh layout, we can extract clothing styles and transfer them to other subjects.}
    \label{fig:garment_switch}
    \vspace{-2mm}
\end{figure}

In our final experiment, we want to demonstrate a potential application of our method, namely garment transfer or virtual try-on.
We take several results of our method and use them to synthesize a subject in new clothing.
To achieve this, we keep the SMPL shape parameters $\shape$.
Then we alter normal and displacement maps according to a different result.
Hereby, we keep details in the facial region, to preserve the subject's identity and hair-style.
Since we edit in UV space, this operation can simply be done using standard image editing techniques.
In Fig.~\ref{fig:garment_switch} we show a subject in three different synthesized clothing styles.

\section{Discussion and Conclusion}
\label{sec:conclusion}
\begin{figure}
    \centering
    \includegraphics[width=\columnwidth]{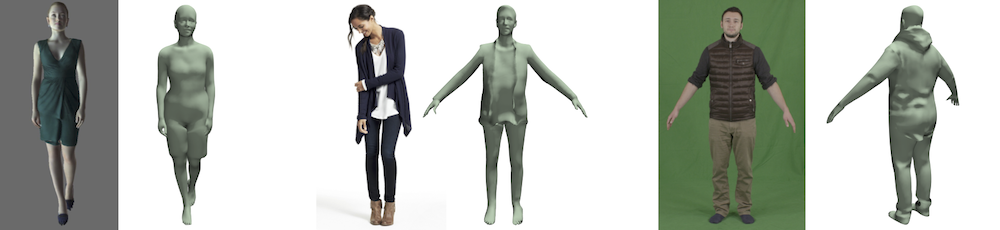}\\
    \caption{Failure cases of our method: The predictor confuses a dress with short pants, a female subject with a male, and hallucinates a hood from a collar.}
    \label{fig:limitations}
    \vspace{-2mm}
\end{figure}

\vspace{-.5mm}
We have proposed a simple yet effective method to infer full-body shape of humans from a single input image.
For the first time, we present single image shape reconstruction with fine details also on occluded parts.
The key idea of this work is to turn a hard full-body shape reconstruction problem into an easier 3D pose-independent image-to-image translation one.
Our model Tex2Shape takes partial texture maps created from DensePose as input and estimates details in the UV-space in form of normal and displacement maps.
The estimated UV maps allow augmenting the SMPL body model with high-frequent details without the need for high mesh resolution.
Our experiments demonstrate that Tex2Shape generalizes robustly to real-world footage, while being trained on synthetic data only.

Our method finds its limitations in hair and clothing that is not covered by the training set.
This is especially the case for long hair and dresses since they cannot be modeled as vector displacement fields.
Typical failure cases are depicted in Fig.~\ref{fig:limitations}.
These failures can be explained with garment-type or gender confusion, caused by missing training samples. In future work, we would like to further open up the problem of human shape estimation and explore shape representations that allow all types of clothing and even accessories.

We have shown, that by transferring a hard problem into a simple formulation, complex models can be outperformed.
Our method lays the foundation for wide-spread 3D reconstruction of people for various applications and even from legacy material.

\hyphenation{For-schungs-ge-mein-schaft}

\noindent
\begin{minipage}{\columnwidth}
    \vspace{2mm}
    \footnotesize
	\noindent
	\textbf{Acknowledgments.}~
	This is work is partly funded by the Deutsche Forschungsgemeinschaft (DFG, German Research Foundation) - 409792180 (Emmy Noether Programme, project: Real Virtual Humans) and project MA2555/12-1. We would like to thank Twindom for providing us with the scan data.
\end{minipage}

{\small
\bibliographystyle{ieee_fullname}
\bibliography{egbib}

\begin{thebibliography}{10}\itemsep=-1pt

\bibitem{code}
\url{http://virtualhumans.mpi-inf.mpg.de/tex2shape/}.

\bibitem{alldieck2019learning}
Thiemo Alldieck, Marcus Magnor, Bharat~Lal Bhatnagar, Christian Theobalt, and
  Gerard Pons-Moll.
\newblock Learning to reconstruct people in clothing from a single {RGB}
  camera.
\newblock In {\em {IEEE} Conf. on Computer Vision and Pattern Recognition},
  2019.

\bibitem{alldieck2018detailed}
Thiemo Alldieck, Marcus Magnor, Weipeng Xu, Christian Theobalt, and Gerard
  Pons-Moll.
\newblock Detailed human avatars from monocular video.
\newblock In {\em International Conf. on 3D Vision}, sep 2018.

\bibitem{alldieck2018video}
Thiemo Alldieck, Marcus Magnor, Weipeng Xu, Christian Theobalt, and Gerard
  Pons-Moll.
\newblock Video based reconstruction of {3D} people models.
\newblock In {\em {IEEE} Conf. on Computer Vision and Pattern Recognition},
  2018.

\bibitem{alp2018densepose}
R{\i}za Alp~G{\"u}ler, Natalia Neverova, and Iasonas Kokkinos.
\newblock Densepose: Dense human pose estimation in the wild.
\newblock In {\em {IEEE} Conf. on Computer Vision and Pattern Recognition},
  pages 7297--7306, 2018.

\bibitem{anguelov2005scape}
Dragomir Anguelov, Praveen Srinivasan, Daphne Koller, Sebastian Thrun, Jim
  Rodgers, and James Davis.
\newblock {SCAPE}: shape completion and animation of people.
\newblock In {\em ACM Transactions on Graphics}, volume~24, pages 408--416.
  ACM, 2005.

\bibitem{bednarik2018learning}
Jan Bednarik, Pascal Fua, and Mathieu Salzmann.
\newblock Learning to reconstruct texture-less deformable surfaces from a
  single view.
\newblock In {\em International Conf. on 3D Vision}, pages 606--615, 2018.

\bibitem{blinn1976texture}
James~F Blinn and Martin~E Newell.
\newblock Texture and reflection in computer generated images.
\newblock {\em Communications of the ACM}, 19(10):542--547, 1976.

\bibitem{Bogo:ICCV:2015}
Federica Bogo, Michael~J. Black, Matthew Loper, and Javier Romero.
\newblock Detailed full-body reconstructions of moving people from monocular
  {RGB-D} sequences.
\newblock In {\em {IEEE} International Conf. on Computer Vision}, pages
  2300--2308, 2015.

\bibitem{bogo2016smplify}
Federica Bogo, Angjoo Kanazawa, Christoph Lassner, Peter Gehler, Javier Romero,
  and Michael~J Black.
\newblock Keep it {SMPL}: Automatic estimation of {3D} human pose and shape
  from a single image.
\newblock In {\em European Conf. on Computer Vision}. Springer, 2016.

\bibitem{bronstein2017geometric}
Michael~M Bronstein, Joan Bruna, Yann LeCun, Arthur Szlam, and Pierre
  Vandergheynst.
\newblock Geometric deep learning: going beyond euclidean data.
\newblock {\em IEEE Signal Processing Magazine}, 2017.

\bibitem{cao2017realtime}
Zhe Cao, Tomas Simon, Shih-En Wei, and Yaser Sheikh.
\newblock Realtime multi-person 2d pose estimation using part affinity fields.
\newblock In {\em {IEEE} Conf. on Computer Vision and Pattern Recognition},
  2017.

\bibitem{danvevrek2017deepgarment}
R Dan{\v{e}}{\v{r}}ek, Endri Dibra, C {\"O}ztireli, Remo Ziegler, and Markus
  Gross.
\newblock Deepgarment: 3d garment shape estimation from a single image.
\newblock In {\em Computer Graphics Forum}, volume~36, pages 269--280. Wiley
  Online Library, 2017.

\bibitem{gardner-sigasia-17}
Marc-Andr\'{e} Gardner, Kalyan Sunkavalli, Ersin Yumer, Xiaohui Shen, Emiliano
  Gambaretto, Christian Gagn\'{e}, and Jean-Fran\c{c}ois Lalonde.
\newblock Learning to predict indoor illumination from a single image.
\newblock {\em ACM Transactions on Graphics}, 9(4), 2017.

\bibitem{guan2009estimating}
Peng Guan, Alexander Weiss, Alexandru~O B{\u{a}}lan, and Michael~J Black.
\newblock Estimating human shape and pose from a single image.
\newblock In {\em {IEEE} International Conf. on Computer Vision}, 2009.

\bibitem{Habermann:2019:LiveCap}
Marc Habermann, Weipeng Xu, Michael Zollh\"{o}fer, Gerard Pons-Moll, and
  Christian Theobalt.
\newblock Livecap: Real-time human performance capture from monocular video.
\newblock {\em ACM Transactions on Graphics}, 38(2):14:1--14:17, 2019.

\bibitem{hasler2009statistical}
Nils Hasler, Carsten Stoll, Martin Sunkel, Bodo Rosenhahn, and H-P Seidel.
\newblock A statistical model of human pose and body shape.
\newblock In {\em Computer Graphics Forum}, 2009.

\bibitem{huynh2018mesoscopic}
Loc Huynh, Weikai Chen, Shunsuke Saito, Jun Xing, Koki Nagano, Andrew Jones,
  Paul Debevec, and Hao Li.
\newblock Mesoscopic facial geometry inference using deep neural networks.
\newblock In {\em Proceedings of the IEEE Conference on Computer Vision and
  Pattern Recognition}, pages 8407--8416, 2018.

\bibitem{insafutdinov2016deepercut}
Eldar Insafutdinov, Leonid Pishchulin, Bjoern Andres, Mykhaylo Andriluka, and
  Bernt Schieke.
\newblock Deepercut: A deeper, stronger, and faster multi-person pose
  estimation model.
\newblock In {\em European Conf. on Computer Vision}, 2016.

\bibitem{isola2017pix2pix}
Phillip Isola, Jun-Yan Zhu, Tinghui Zhou, and Alexei~A Efros.
\newblock Image-to-image translation with conditional adversarial networks.
\newblock In {\em {IEEE} Conf. on Computer Vision and Pattern Recognition},
  pages 1125--1134, 2017.

\bibitem{jackson20183d}
Aaron~S Jackson, Chris Manafas, and Georgios Tzimiropoulos.
\newblock 3d human body reconstruction from a single image via volumetric
  regression.
\newblock In {\em European Conference on Computer Vision}, pages 64--77.
  Springer, 2018.

\bibitem{jain2010moviereshape}
Arjun Jain, Thorsten Thorm{\"a}hlen, Hans-Peter Seidel, and Christian Theobalt.
\newblock Moviereshape: Tracking and reshaping of humans in videos.
\newblock In {\em ACM Transactions on Graphics}, volume~29, page 148. ACM,
  2010.

\bibitem{jin2018pixel}
Ning Jin, Yilin Zhu, Zhenglin Geng, and Ronald Fedkiw.
\newblock A pixel-based framework for data-driven clothing.
\newblock {\em arXiv preprint arXiv:1812.01677}, 2018.

\bibitem{joo2018total}
Hanbyul Joo, Tomas Simon, and Yaser Sheikh.
\newblock Total capture: A 3d deformation model for tracking faces, hands, and
  bodies.
\newblock In {\em {IEEE} Conf. on Computer Vision and Pattern Recognition},
  pages 8320--8329, 2018.

\bibitem{kanamori2018relighting}
Yoshihiro Kanamori and Yuki Endo.
\newblock Relighting humans: occlusion-aware inverse rendering for fullbody
  human images.
\newblock {\em ACM Transactions on Graphics}, 37(270):1--270, 2018.

\bibitem{kanazawa2018endtoend}
Angjoo Kanazawa, Michael~J. Black, David~W. Jacobs, and Jitendra Malik.
\newblock End-to-end recovery of human shape and pose.
\newblock In {\em {IEEE} Conf. on Computer Vision and Pattern Recognition},
  2018.

\bibitem{kingma2014adam}
Diederik~P Kingma and Jimmy Ba.
\newblock Adam: A method for stochastic optimization.
\newblock In {\em International Conference on Learning Representations},
  volume~5, 2015.

\bibitem{lahner2018deepwrinkles}
Zorah Lahner, Daniel Cremers, and Tony Tung.
\newblock Deepwrinkles: Accurate and realistic clothing modeling.
\newblock In {\em European Conf. on Computer Vision}, pages 667--684, 2018.

\bibitem{Lassner}
Christoph Lassner, Javier Romero, Martin Kiefel, Federica Bogo, Michael~J
  Black, and Peter~V Gehler.
\newblock Unite the people: Closing the loop between 3d and 2d human
  representations.
\newblock In {\em {IEEE} Conf. on Computer Vision and Pattern Recognition},
  2017.

\bibitem{LWSLVDT13}
Guannan Li, Chenglei Wu, Carsten Stoll, Yebin Liu, Kiran Varanasi, Qionghai
  Dai, and Christian Theobalt.
\newblock Capturing relightable human performances under general uncontrolled
  illumination.
\newblock In {\em Computer Graphics Forum}, 2013.

\bibitem{liuLQWTcvpr16DeepFashion}
Ziwei Liu, Ping Luo, Shi Qiu, Xiaogang Wang, and Xiaoou Tang.
\newblock Deepfashion: Powering robust clothes recognition and retrieval with
  rich annotations.
\newblock In {\em {IEEE} Conf. on Computer Vision and Pattern Recognition},
  2016.

\bibitem{liuYLWTeccv16FashionLandmark}
Ziwei Liu, Sijie Yan, Ping Luo, Xiaogang Wang, and Xiaoou Tang.
\newblock Fashion landmark detection in the wild.
\newblock In {\em European Conf. on Computer Vision}, 2016.

\bibitem{smpl2015loper}
Matthew Loper, Naureen Mahmood, Javier Romero, Gerard Pons-Moll, and Michael~J
  Black.
\newblock {SMPL}: A skinned multi-person linear model.
\newblock {\em ACM Transactions on Graphics}, 2015.

\bibitem{natsume2018siclope}
Ryota Natsume, Shunsuke Saito, Zeng Huang, Weikai Chen, Chongyang Ma, Hao Li,
  and Shigeo Morishima.
\newblock Siclope: Silhouette-based clothed people.
\newblock In {\em {IEEE} Conf. on Computer Vision and Pattern Recognition},
  2019.

\bibitem{neverova2018dense}
Natalia Neverova, Riza Alp~Guler, and Iasonas Kokkinos.
\newblock Dense pose transfer.
\newblock In {\em European Conf. on Computer Vision}, 2018.

\bibitem{omran2018neural}
Mohamed Omran, Christop Lassner, Gerard Pons-Moll, Peter Gehler, and Bernt
  Schiele.
\newblock Neural body fitting: Unifying deep learning and model based human
  pose and shape estimation.
\newblock In {\em International Conf. on 3D Vision}, 2018.

\bibitem{pavlakos2018humanshape}
Georgios Pavlakos, Luyang Zhu, Xiaowei Zhou, and Kostas Daniilidis.
\newblock Learning to estimate 3{D} human pose and shape from a single color
  image.
\newblock In {\em {IEEE} Conf. on Computer Vision and Pattern Recognition},
  2018.

\bibitem{pishchulin16cvpr}
Leonid Pishchulin, Eldar Insafutdinov, Siyu Tang, Bjoern Andres, Mykhaylo
  Andriluka, Peter Gehler, and Bernt Schiele.
\newblock Deepcut: Joint subset partition and labeling for multi person pose
  estimation.
\newblock In {\em {IEEE} Conf. on Computer Vision and Pattern Recognition},
  2016.

\bibitem{ponsmoll2017clothcap}
Gerard Pons-Moll, Sergi Pujades, Sonny Hu, and Michael Black.
\newblock {ClothCap}: Seamless {4D} clothing capture and retargeting.
\newblock {\em ACM Transactions on Graphics}, 36(4), 2017.

\bibitem{pons2015dyna}
Gerard Pons-Moll, Javier Romero, Naureen Mahmood, and Michael~J Black.
\newblock Dyna: a model of dynamic human shape in motion.
\newblock {\em ACM Transactions on Graphics}, 34:120, 2015.

\bibitem{popa2009wrinkling}
Tiberiu Popa, Quan Zhou, Derek Bradley, Vladislav Kraevoy, Hongbo Fu, Alla
  Sheffer, and Wolfgang Heidrich.
\newblock Wrinkling captured garments using space-time data-driven deformation.
\newblock In {\em Computer Graphics Forum}, volume~28, pages 427--435, 2009.

\bibitem{ramamoorthi2001efficient}
Ravi Ramamoorthi and Pat Hanrahan.
\newblock An efficient representation for irradiance environment maps.
\newblock In {\em Proceedings of the 28th Annual Conference on Computer
  Graphics and Interactive Techniques}, pages 497--500. ACM, 2001.

\bibitem{rogge2014garment}
Lorenz Rogge, Felix Klose, Michael Stengel, Martin Eisemann, and Marcus Magnor.
\newblock Garment replacement in monocular video sequences.
\newblock {\em ACM Transactions on Graphics}, 34(1):6, 2014.

\bibitem{sela2017unrestricted}
Matan Sela, Elad Richardson, and Ron Kimmel.
\newblock Unrestricted facial geometry reconstruction using image-to-image
  translation.
\newblock In {\em {IEEE} Conf. on Computer Vision and Pattern Recognition},
  pages 1576--1585, 2017.

\bibitem{sengupta2018sfsnet}
Soumyadip Sengupta, Angjoo Kanazawa, Carlos~D Castillo, and David~W Jacobs.
\newblock Sfsnet: Learning shape, reflectance and illuminance of facesin the
  wild'.
\newblock In {\em {IEEE} Conf. on Computer Vision and Pattern Recognition},
  pages 6296--6305, 2018.

\bibitem{SimulCap19}
Yu Tao, Zerong Zheng, Yuan Zhong, Jianhui Zhao, Dai Quionhai, Gerard Pons-Moll,
  and Yebin Liu.
\newblock Simulcap : Single-view human performance capture with cloth
  simulation.
\newblock In {\em {IEEE} Conf. on Computer Vision and Pattern Recognition}, jun
  2019.

\bibitem{FML2019}
Ayush Tewari, Florian Bernard, Pablo Garrido, Gaurav Bharaj, Mohammed Elgharib,
  Hans-Peter Seidel, Patrick Perez, Michael Zollh{\"o}fer, and Christian
  Theobalt.
\newblock Fml: Face model learning from videos.
\newblock In {\em {IEEE} Conf. on Computer Vision and Pattern Recognition},
  2019.

\bibitem{tewari2018self}
Ayush Tewari, Michael Zollh{\"o}fer, Pablo Garrido, Florian Bernard, Hyeongwoo
  Kim, Patrick P{\'e}rez, and Christian Theobalt.
\newblock Self-supervised multi-level face model learning for monocular
  reconstruction at over 250 hz.
\newblock In {\em {IEEE} Conf. on Computer Vision and Pattern Recognition},
  2018.

\bibitem{tung2017self}
Hsiao-Yu Tung, Hsiao-Wei Tung, Ersin Yumer, and Katerina Fragkiadaki.
\newblock Self-supervised learning of motion capture.
\newblock In {\em Advances in Neural Information Processing Systems}, pages
  5236--5246, 2017.

\bibitem{varol2018bodynet}
G{\"u}l Varol, Duygu Ceylan, Bryan Russell, Jimei Yang, Ersin Yumer, Ivan
  Laptev, and Cordelia Schmid.
\newblock Bodynet: Volumetric inference of 3d human body shapes.
\newblock In {\em European Conf. on Computer Vision}, 2018.

\bibitem{varol17_surreal}
G{\"u}l Varol, Javier Romero, Xavier Martin, Naureen Mahmood, Michael~J. Black,
  Ivan Laptev, and Cordelia Schmid.
\newblock Learning from synthetic humans.
\newblock In {\em {IEEE} Conf. on Computer Vision and Pattern Recognition},
  2017.

\bibitem{vonMarcard2018}
Timo von Marcard, Roberto Henschel, Michael Black, Bodo Rosenhahn, and Gerard
  Pons-Moll.
\newblock Recovering accurate 3d human pose in the wild using imus and a moving
  camera.
\newblock In {\em European Conf. on Computer Vision}, sep 2018.

\bibitem{wang2018learning}
Tuanfeng~Y. Wang, Duygu Ceylan, Jovan Popovic, and Niloy~J. Mitra.
\newblock Learning a shared shape space for multimodal garment design.
\newblock {\em ACM Transactions on Graphics}, 37(6):1:1--1:14, 2018.

\bibitem{wang2003multiscale}
Zhou Wang, Eero~P Simoncelli, and Alan~C Bovik.
\newblock Multiscale structural similarity for image quality assessment.
\newblock In {\em Asilomar Conference on Signals, Systems \& Computers},
  volume~2, pages 1398--1402, 2003.

\bibitem{weng2018photo}
Chung-Yi Weng, Brian Curless, and Ira Kemelmacher-Shlizerman.
\newblock Photo wake-up: 3d character animation from a single photo.
\newblock In {\em {IEEE} Conf. on Computer Vision and Pattern Recognition},
  2019.

\bibitem{wu2013set}
Chenglei Wu, Carsten Stoll, Levi Valgaerts, and Christian Theobalt.
\newblock On-set performance capture of multiple actors with a stereo camera.
\newblock {\em ACM Transactions on Graphics}, 32(6):161, 2013.

\bibitem{Wu:2012}
Chenglei Wu, Kiran Varanasi, and Christian Theobalt.
\newblock Full body performance capture under uncontrolled and varying
  illumination: A shading-based approach.
\newblock In {\em European Conf. on Computer Vision}, pages 757--770, 2012.

\bibitem{MonoPerfCap_SIGGRAPH2018}
Weipeng Xu, Avishek Chatterjee, Michael Zollhoefer, Helge Rhodin, Dushyant
  Mehta, Hans-Peter Seidel, and Christian Theobalt.
\newblock Monoperfcap: Human performance capture from monocular video.
\newblock {\em ACM Transactions on Graphics}, 2018.

\bibitem{yang2018analyzing}
Jinlong Yang, Jean-S{\'e}bastien Franco, Franck H{\'e}troy-Wheeler, and
  Stefanie Wuhrer.
\newblock Analyzing clothing layer deformation statistics of 3d human motions.
\newblock In {\em European Conf. on Computer Vision}, pages 237--253, 2018.

\bibitem{zahng2017shapeundercloth}
Chao Zhang, Sergi Pujades, Michael Black, and Gerard Pons-Moll.
\newblock Detailed, accurate, human shape estimation from clothed {3D} scan
  sequences.
\newblock In {\em {IEEE} Conf. on Computer Vision and Pattern Recognition},
  2017.

\bibitem{zhang1999shape}
Ruo Zhang, Ping-Sing Tsai, James~Edwin Cryer, and Mubarak Shah.
\newblock Shape-from-shading: a survey.
\newblock {\em IEEE Transactions on Pattern Analysis and Machine Intelligence},
  21(8):690--706, 1999.

\bibitem{Zheng2019DeepHuman}
Zerong Zheng, Tao Yu, Yixuan Wei, Qionghai Dai, and Yebin Liu.
\newblock Deephuman: 3d human reconstruction from a single image.
\newblock {\em arXiv preprint arXiv:1903.06473}, Sept 2019.

\bibitem{zhou2010parametric}
Shizhe Zhou, Hongbo Fu, Ligang Liu, Daniel Cohen-Or, and Xiaoguang Han.
\newblock Parametric reshaping of human bodies in images.
\newblock In {\em ACM Transactions on Graphics}, volume~29, page 126. ACM,
  2010.

\end{thebibliography}
}

\appendix
\newpage
\section{Appendix}

We show here additional experiments to understand the influence of illumination on our model and its robustness to varying camera intrinsics.
We evaluate the  $\beta$-regression network and perform an ablation of the UV map resolution.
Finally, we present more qualitative results.

\subsection{Influence of Illumination}
As already emphasized in the main paper, shading is potentially a strong cue for our model.
In the following, we evaluate the illumination augmentation during training and the robustness of our model to varying illumination.

\begin{figure*}
    \centering
    \includegraphics[width=\textwidth]{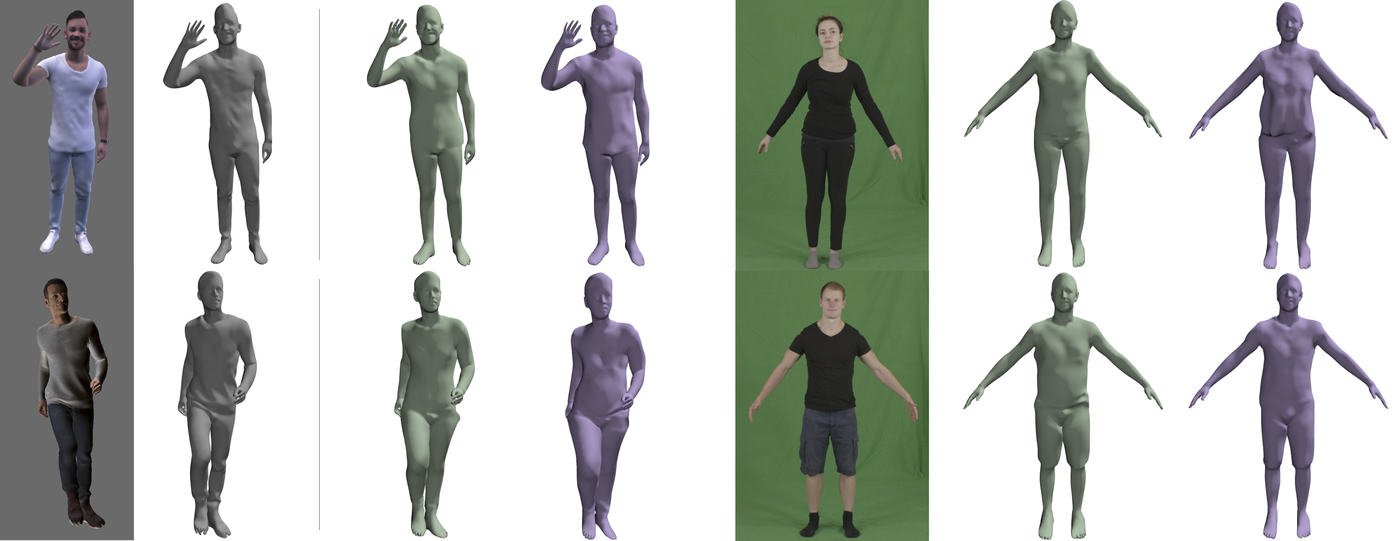}\
    \caption{Our method (green) compared to our method trained without illumination augmentation (purple) and ground truth (grey). Looking closely, we notice worse performance specially on the face region, and artifacts for the method without illumination augmentation. Notice for example the example on the bottom left, the face, legs shape, and chest region is more accurately reconstructed when using augmentation (green). }
    \label{fig:no_shading}

    \vspace{5mm}
    \includegraphics[width=\textwidth]{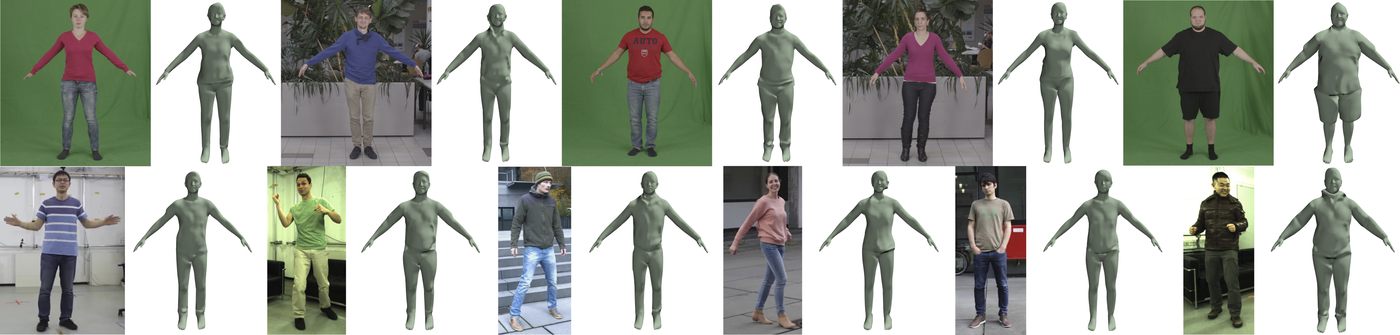}\\
    \caption{3D reconstruction results on two in-the-wild datasets: PeopleSnapshot~\cite{alldieck2018video} (1st row) and MonoPerfCap~\cite{MonoPerfCap_SIGGRAPH2018} (2nd row).}
    \label{fig:qualitative_results}
    
\end{figure*}

In order to evaluate the effect of the illumination augmentation during training, we re-trained our model with constant ambient illumination. 
This means we render the scans using the textures only.
While being scanned, the subjects have been exposed to uniform lighting.
However, shading is still present in wrinkles and smaller structures.
This means, we cannot factor out shading effects completely.
Nevertheless, in Fig.~\ref{fig:no_shading} we can see more consistent details for our final method, especially for the faces.

\begin{figure}
	\centering
	\includegraphics[width=\columnwidth]{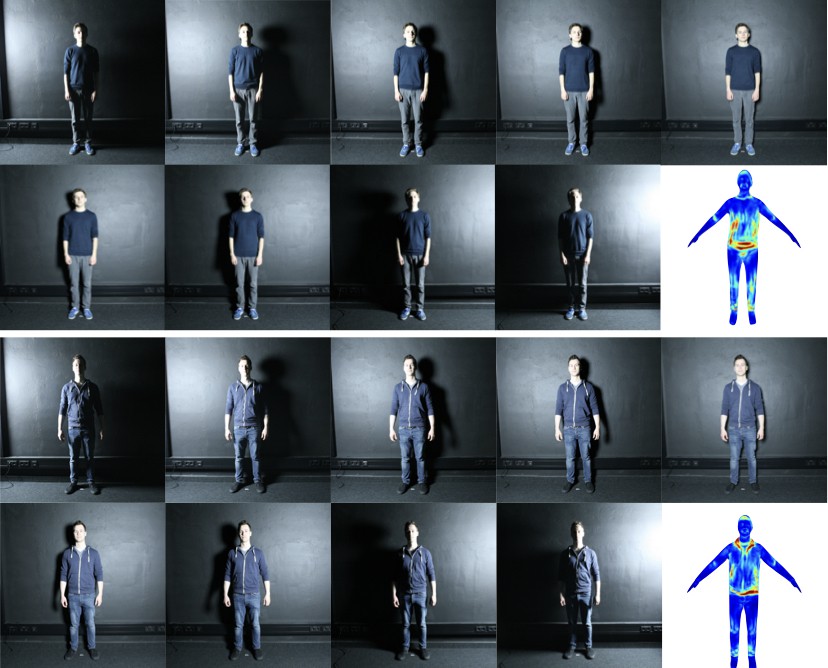}\\
	\caption{Displacement reconstruction consistency under varying illumination. The heatmap illustrates the vector norm of per surface point standard deviation  (dark-red means $\geq4$cm).}
	\label{fig:illu_std}
\end{figure}

Our model should produce the same or at least a very similar result when applied on two different photos of the same person in the same clothing but under varying illumination.
To validate illumination invariance of our model, we took 9 photos of two subjects while rotating the light-source around the subject.
In Fig.~\ref{fig:illu_std} we show the different photos and a heat-map illustrating areas with high standard deviation.
We see a consistent picture with varying details only in areas of likely fabric movement.

\subsection{Influence of Camera Intrinsics}
\begin{figure}
	\centering
	\includegraphics[width=\columnwidth]{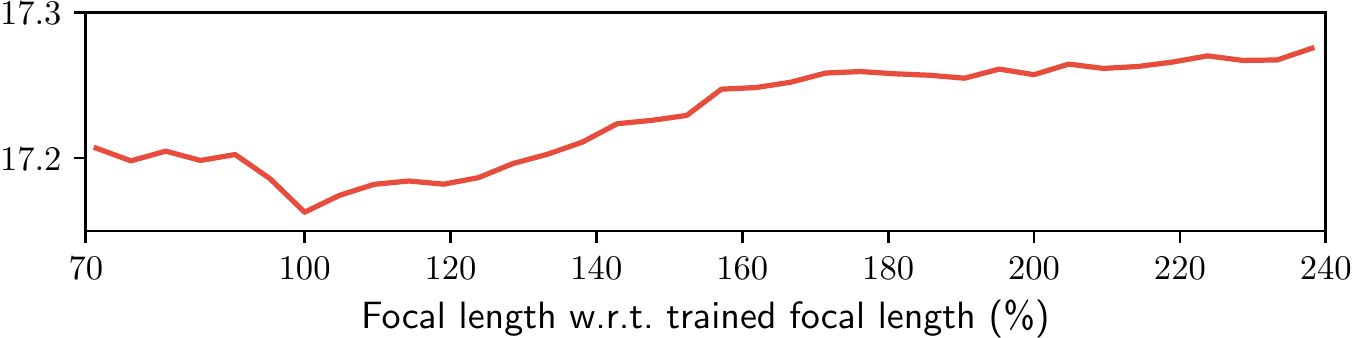}\\
	\caption{Mean SMPL vertex-to-vertex error in mm (without added displacements) over the test-set for varying focal length.}
	\label{fig:intrinsics_vs_accuracy}
\end{figure}

Camera intrinsics are mostly unknown at test time, especially for in-the-wild photos.
The focal length is an important camera parameter, which can affect the results of our method.
We have trained our model with a fixed focal length.
To study the robustness of our method against varying focal length, we render our test set in A-poses with different focal length and distance to the camera.
We keep the ratio between distance and focal length fixed, creating a \emph{Vertigo Effect}.
In Fig.~\ref{fig:intrinsics_vs_accuracy}, we report the mean vertex-to-vertex error of the naked SMPL model under varying focal length.
Although the lowest error is obtained for the focal length assumed during training, different focal lengths increase the error only slightly, which demonstrates the robustness of our model.

\subsection{Numerical Comparison with HMR}
In order to evaluate the $\beta$-regression network, we compare our naked results without added displacements against HMR \cite{kanazawa2018endtoend}.
Since we do not estimate pose it has to be factored out before comparison.
To this end, we follow the established procedure in \cite{Bogo:ICCV:2015} and adjust pose and scale
of the results of both methods to match the ground truth scans.
On our test-set, our method using DensePose mapping achieves a mean bi-directional vertex to surface error of $10.57 \pm10.68$mm compared to the clothed scans.
HMR achieves $16.28\pm 17.05$mm.
Our method can better estimate the body shapes.
This is likely linked to the fact, that our method directly uses dense image-space detections, while HMR correlates surface with bone-lengths.
With added displacements, our method achieves $5.19\pm 6.36$mm.
All results are up to scale.

\subsection{UV Resolution Ablation}
\begin{figure}
	\centering
	\includegraphics[width=\columnwidth]{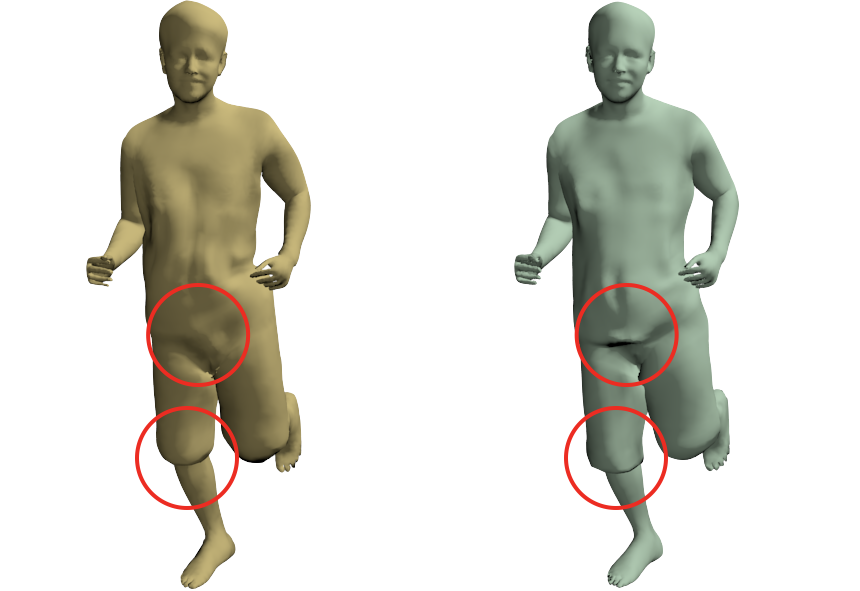}\\
	\caption{Comparison of two variants of our network: Using $256\times256$px resolution (left) decreased the quality only slightly when compared to the original resolution of $512\times512$px (right).}
	\label{fig:256_ablation}
	
\end{figure}

To evaluate our choice of the UV resolution ($512\times512$px), we train a variant of the network with $256\times256$px maps.
The results look surprisingly good.
A close inspection of the results reveals missing details and smoothed edges.
An example is shown in Fig.~\ref{fig:256_ablation}.
However, this experiment demonstrates that Tex2Shape can be trained with lower resolution without largely decreased quality.

\subsection{Additional Qualitative Results}
In Fig.~\ref{fig:qualitative_results}, we show more in-the-wild results of our method on \emph{MonoPerfCap}~\cite{MonoPerfCap_SIGGRAPH2018} and \emph{PeopleSnapshot}~\cite{alldieck2018video} datasets. 

\newpage


\vfill\null

\end{document}